\documentclass[11pt]{article}

\usepackage[final]{acl}

\usepackage{times}
\usepackage{latexsym}
\usepackage[T1]{fontenc}
\usepackage[utf8]{inputenc}
\usepackage{microtype}
\usepackage{inconsolata}
\usepackage{booktabs}
\usepackage{tabularx}
\usepackage{array}
\usepackage{graphicx}

\title{ProcessPrefBench: Evaluating Cross-Task Transfer of Individual Process Preferences from Naturalistic Human--AI Interaction}

\author{Yiwen Yang \\
  Independent Researcher \\
  \texttt{getupyang@gmail.com} \\\And
  Guanming Liu \\
  Fudan University \\
  \texttt{gmliu24@m.fudan.edu.cn}}

\begin{document}
\maketitle

\begin{abstract}
In natural human--AI dialogue, users' corrections, requests, and self-descriptions carry reusable signals about how they want work to be done. These signals reveal \emph{individual process preferences}, user-specific standards for evidence, depth, analytical focus, procedure, and delivery. We ask whether a preference revealed in one dialogue can shape an assistant's later answer when the task does not restate it. ProcessPrefBench uses three months of natural dialogues from two knowledge workers to build 66 source--target cases separately judged for source evidence and target applicability. Sources are preselected, so the benchmark tests application rather than retrieval. We test three models with target-only, full-history, and distilled-preference prompts. Target-specific 0--2 rubrics score preference realization, and a task gate flags severe failures. Compared with target-only prompting, full-history and imperative distilled-preference prompting raise preference realization by 0.204 and 0.244 points on the 0--2 scale, respectively, with no reliable advantage from distillation. Fabricated-evidence failures rise from 5.8\% under target-only prompting to 10.1\% under full-history and 10.4\% under imperative prompting. ProcessPrefBench highlights a tension in learning from dialogue, where drawing on prior interactions can improve preference realization yet also invite fabricated evidence. We hope it helps build assistants that honor a user's process preferences without sacrificing factual reliability.
\end{abstract}

\section{Introduction}

\begin{figure*}[t]
\centering
\includegraphics[width=\textwidth]{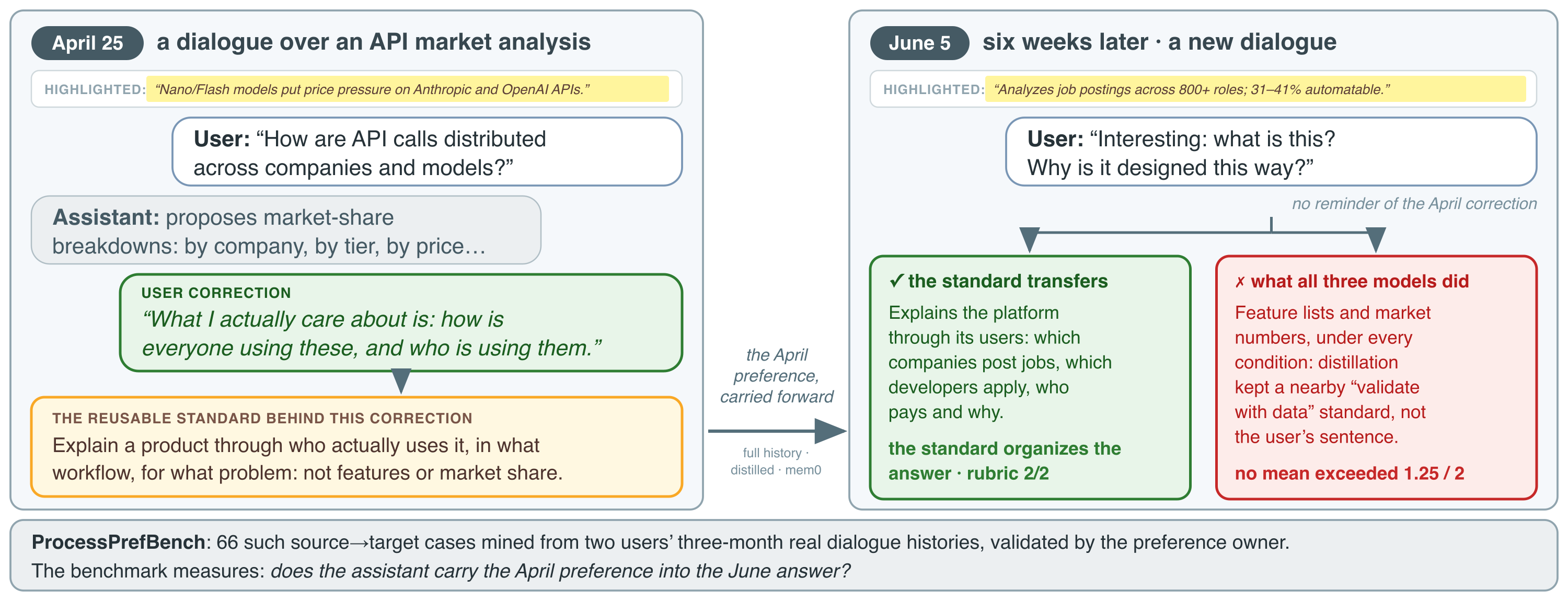}
\caption{A ProcessPrefBench case (translated from Chinese). An April dialogue establishes a reusable analysis standard; six weeks later, the same user asks about a different product in a new dialogue without restating it. Appendix~\ref{app:rubric} audits this case end to end.}
\label{fig:example}
\end{figure*}

Users often do not state their process preferences in advance. In human--AI dialogue, these preferences become visible when users react to a particular answer through corrections, follow-up requests, or self-descriptions. Remembering facts about a user does not by itself teach an assistant how that user wants work to be carried out. Figure~\ref{fig:example} follows an April correction that redirects product analysis from market-share breakdowns toward who uses a product and how. Six weeks later, the same user asks about a different product without restating that standard. We define individual process preferences as reusable, user-specific standards for how work should be done. They may apply beyond the dialogue in which they are expressed and govern what evidence supports a claim, how an analysis is organized, how deeply a source is explained, and whether a requested artifact is delivered without further confirmation.

Learning process preferences from dialogue is a delayed, cross-task form of adaptation. An end-to-end system would have to infer the reusable standard from the dialogue that carries it, determine when it applies, and realize it without treating remembered content as evidence for the new task. These stages can fail in different ways. The system may preserve the wrong standard, fail to realize a correctly preserved one, or import remembered content as unsupported target claims. ProcessPrefBench studies a narrower setting in which the relevant source dialogue is preselected for each target. It tests whether models can realize the preference when given the full dialogue or a distilled representation without importing unsupported source content.

Existing benchmarks approach learning from dialogue in two ways: some test whether conversational information remains available for later reasoning~\citep{maharana2024locomo,wu2025longmemeval}, while others hand the model a profile or preference record directly~\citep{jiang2025personamem,zhao2025prefeval}. These settings do not jointly test whether a preference can be induced from a natural dialogue and then applied to a different task without importing unsupported content.

We introduce ProcessPrefBench using three months of natural user--assistant dialogues from Margin, a reading, annotation, and AI-discussion product used primarily by knowledge workers. The benchmark is an idiographic diagnostic instrument focused on research, sensemaking, and decision tasks from two users. Each case links a preference-bearing source to a different task from the same user; construction separately establishes that the source supports a reusable standard and that the standard should affect a target that does not restate it.

ProcessPrefBench contributes a source-to-target construction protocol, target-specific rubrics that distinguish surface mention from functional integration, and an evaluation that separates preference realization from severe task failure. The relevant source is preselected for every case, so the experiment tests application of a known-relevant dialogue rather than retrieval from a heterogeneous history. Across three models, relevant history improves preference realization, but distillation offers no reliable advantage over full history and both memory conditions increase fabricated-evidence failures.

\section{Related Work}

Long-term-memory benchmarks test whether assistants retain and reason over information distributed across extended interactions. LoCoMo and LongMemEval cover extraction, temporal reasoning, and knowledge updates across extended conversations, while MemoryAgentBench uses incremental multi-turn interactions~\citep{maharana2024locomo,wu2025longmemeval,hu2025memoryagentbench}. These benchmarks establish whether information remains available over time; ProcessPrefBench asks whether a dialogue also supports a reusable working standard.

Personalization benchmarks typically provide the user signal more directly. PersonaMem studies dynamic user profiling, LaMP evaluates personalized language-model tasks, and PrefEval tests preference recognition and following~\citep{jiang2025personamem,salemi2024lamp,zhao2025prefeval}. CUPID infers which contextual preference applies to a new request in constructed interactions~\citep{kim2025cupid}, while AlpsBench builds personalization tests from natural dialogues~\citep{xiao2026alpsbench}. ProcessPrefBench instead derives each standard from a natural correction, prescription, or self-description and establishes its applicability to a later target.

Learning preferences from interaction is a closer setting. PRELUDE learns latent preferences from user edits~\citep{gao2024prelude}, and TRACE compiles corrections into runtime enforcement rules~\citep{zhou2026trace}; our focus is the evaluation problem of tracing a source-supported standard into a different task and distinguishing functional realization from surface mention. Game-based evaluation measures interactive ability through situated behavior rather than similarity to reference text~\citep{chalamalasetti2023clembench}; our rubrics apply the same principle to naturalistic collaboration by scoring functional integration instead of surface mention.

External memory systems such as MemoryBank, MemoryOS, and mem0 store and reintroduce information outside the base model~\citep{zhong2024memorybank,kang2025memoryos,chhikara2025mem0}; HaluMem studies hallucinations introduced by memory operations~\citep{chen2025halumem}. A retrieved interaction can mix a reusable process standard with earlier-task content, so retrieval alone neither ensures transfer nor prevents that content from entering target-task claims.

\section{Benchmark Construction}

\paragraph{Data source.}
We collect three months of natural interactions from two knowledge workers using Margin, a webpage reading, annotation, and AI-discussion product. Each trace retains the available reading context, user request, assistant response, and subsequent user turns. Assistant responses came from locally installed Claude Code and Codex at their then-default models; early traces lack per-turn version logs, so the source model is not always identifiable. For 13 cases with truncated reading context, we restore verified source material so that the evaluation does not force a choice between fabrication and refusal.

\paragraph{Human source annotation.}
A source is retained when the user's words support a reusable standard for future work, expressed through correction, prescription, or self-description. Each user annotates their own interactions and excludes factual repairs, current-task-only requests, and scene-bound conclusions because they do not justify transfer beyond the source task. Each retained case stores the inferred standard, its supporting words, and the full interaction as auditable evidence.

\paragraph{Human target pairing.}
Source annotation does not determine a valid target. For each source, the preference owner separately judges another same-user task. A pairing is retained when the target does not repeat the preference but applying the source-supported standard should improve the answer. Review also records order, cue strength, material and tool availability, and evaluability. This procedure rejects topical similarity alone and prevents source validity from making target applicability automatic.

\paragraph{Rubric construction.}
Each case contains one to three source-anchored \emph{preference atoms} with target-specific criteria for not applicable, 0, 1, or 2. A criterion is not applicable when its trigger is absent from the target. A score of 0 marks an absent or contradicted preference, 1 marks local or cosmetic realization, and 2 requires a functional change to the answer's reasoning, organization, conclusion, or delivery. This progression measures whether the remembered standard changes the answer rather than whether the answer merely mentions it. One AI research assistant drafted the target-specific criteria under a frozen specification, followed by owner spot checks. Appendix~\ref{app:rubric} gives two end-to-end case audits.

A separate task gate marks severe omission, major factual error, fabricated evidence, hard-constraint violation, or a false claim of external execution. Task failure does not overwrite the preference score: keeping the two judgments separate prevents a personalized but unusable answer from counting as success and prevents lack of personalization alone from becoming a general task failure. Among these failure types, fabricated evidence carries our second headline result. The gate assigns this label when an answer presents specific claims about papers, people, data, citations, or links as established evidence that the target-side materials do not support, or falsely claims an external action. The label is operational. Because the gate does not see the source, a carried-over detail and an invented one receive the same label when neither has target-side support.

\paragraph{Measurement audit.}
The benchmark changed in response to three measurement checks. An ablation supported removing descriptive reference answers that did not clarify the target-specific rubric. A controlled comparison exposed wording that favored injected preferences and led us to rewrite the affected criteria. A material audit identified truncated reading context that could induce fabrication, motivating the verified restoration described above.

\section{Evaluation}

\paragraph{Conditions.}
All conditions share the system prompt, page title, highlighted context, verified supplied material when needed, and verbatim user question. Target-only prompting adds no memory, while full-history prompting adds the complete source interaction. For both distilled-preference conditions, the tested model distills each source once per source--model pair and reuses the result. Imperative prompting always applies it, whereas conditional prompting applies it only when relevant. The mem0 external-memory system extracts from the source and retrieves against the target. Conditional prompting is our variant rather than a description of mem0 semantics.

\paragraph{Models and generation.}
We test DeepSeek V4 Flash, Qwen-Max, and GPT-4o. Every case is generated twice for each model and memory condition. The final run contains 1980 answers.

\paragraph{Judging.}
Claude Sonnet 4.6 serves in isolated preference and task-gate roles; its model family is disjoint from the tested models, which limits self-preference bias~\citep{zheng2023mtbench}. Both roles see the target, materials, and answer, but not the tested model, condition, source interaction, or alternatives. The preference judge also sees the rubric and casts three evidence-citing votes per atom. A majority of not-applicable votes excludes the atom; otherwise we use the median applicable score. The task gate sees validity checks without the preference rubric and uses two votes plus a third on ties.

\paragraph{Judge validation.}\label{sec:judge-validation}
We audited the judge on 30 answers, oversampling judge-flagged ones; two annotators rated every answer blind to the tested model, condition, and judge scores. Post-calibration human agreement is substantial ($\kappa=.72$ on preference atoms, $.84$ on the task gate), while agreement with the judge is lower and the judge scores preferences more conservatively. The audit is diagnostic rather than confirmatory, so the main effects remain estimates under the benchmark judge; Appendix~\ref{app:human} reports the protocol, full agreement table, and per-annotator ratings.

\paragraph{Aggregation.}
An answer score averages its applicable atoms; model--condition means exclude answers with no applicable atoms, and task failure is reported independently. Primary comparisons contrast imperative distilled-preference prompting with target-only and full-history prompting: we macro-average paired differences across models with exact sign tests and report case-level cluster-bootstrap 95\% confidence intervals (10{,}000 resamples). Conditional distilled-preference prompting is descriptive because every formal case is an intended positive match.

\section{Results}

\begin{table*}[t]
\centering
\small
\setlength{\tabcolsep}{4.2pt}
\begin{tabularx}{\textwidth}{@{}Xcccccc@{}}
\toprule
& \multicolumn{2}{c}{DeepSeek V4 Flash} & \multicolumn{2}{c}{Qwen-Max} & \multicolumn{2}{c}{GPT-4o} \\
\cmidrule(lr){2-3}\cmidrule(lr){4-5}\cmidrule(lr){6-7}
Condition & \shortstack{Preference\\score} & \shortstack{Task failure\\rate} & \shortstack{Preference\\score} & \shortstack{Task failure\\rate} & \shortstack{Preference\\score} & \shortstack{Task failure\\rate} \\
\midrule
Target-only prompting & 0.79 & 8.3\% & 0.62 & 13.6\% & 0.47 & 23.5\% \\
Full-history prompting & 1.15 & 9.1\% & 0.77 & 23.5\% & 0.59 & 14.4\% \\
Imperative distilled-preference prompting & 1.23 & 14.4\% & 0.73 & 22.7\% & 0.66 & 19.7\% \\
Conditional distilled-preference prompting & 1.11 & 8.3\% & 0.61 & 18.9\% & 0.58 & 23.5\% \\
mem0 external-memory system & 1.06 & 15.2\% & 0.59 & 21.3\% & 0.49 & 22.6\% \\
\bottomrule
\end{tabularx}
\caption{Preference score (higher is better) and task-failure rate (lower is better).}

\label{tab:results}
\end{table*}

\begin{figure}[t]
\centering
\includegraphics[width=.84\columnwidth]{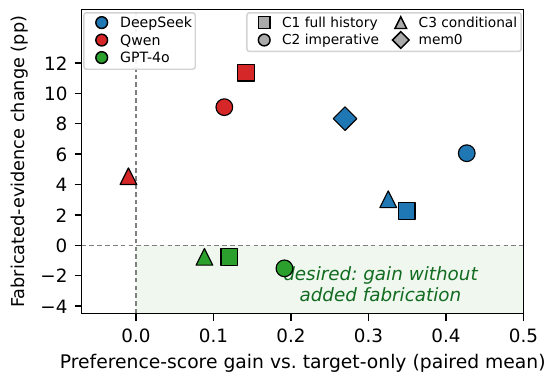}
\caption{Preference-score and fabricated-evidence changes from target-only prompting across model--condition pairs.}
\label{fig:tradeoff}
\end{figure}

\paragraph{Preference transfer.}
Access to user-specific history improves preference scores over target-only prompting. Across models, imperative distilled-preference prompting exceeds target-only prompting by 0.244 points (95\% CI $[0.177, 0.313]$; 48/8/9 wins/ties/losses; exact sign $p<.0001$), with positive effects for DeepSeek V4 Flash (+0.427), Qwen-Max (+0.114), and GPT-4o (+0.192). Full-history prompting gains 0.204 points over target-only prompting (95\% CI $[0.143, 0.269]$; 44/12/9; $p<.0001$).

Distillation does not yield a reliable preference-score advantage over full history. The difference is +0.026 (95\% CI $[-0.034, +0.087]$; $p=.451$), with model-level effects of +0.080 for DeepSeek V4 Flash, $-$0.058 for Qwen-Max, and +0.071 for GPT-4o.\footnote{The gains of 0.244 and 0.204 are computed over the 65 cases with an applicable target-only score, so their difference is 0.040 on that common subset. The direct imperative-versus-full-history comparison includes one additional case for which all target-only answers had no applicable rubric atom; its case-level imperative-versus-full-history difference is $-0.875$, reducing the 66-case mean to 0.026.} This does not establish equivalence; the representations' relative preference effects remain unresolved. The mem0 cell means exceed target-only prompting for DeepSeek V4 Flash and GPT-4o but not for Qwen-Max.

\paragraph{Task validity and factual reliability.}
A preference-score gain does not guarantee answer validity (Figure~\ref{fig:tradeoff}). Imperative distilled-preference prompting raises the overall task-failure rate from 15.2\% to 18.9\% relative to target-only prompting, although the difference is not conventionally significant (exact McNemar $p=.086$). Fabricated-evidence failures rise from 5.8\% under target-only prompting to 10.4\% under imperative prompting, a difference of 4.5 percentage points (95\% CI $[+2.0, +7.6]$; exact McNemar $p=.006$). Full-history prompting shows a similar increase, from 5.8\% to 10.1\% (difference 4.3 points; 95\% CI $[+1.0, +7.8]$; $p=.010$), while the two memory conditions differ by only 0.3 points ($p=1.0$). The increase is largest for Qwen-Max under imperative distilled-preference prompting (+9.1 points, $p=.004$), while GPT-4o shows no same-direction increase. The plotted mem0 comparison also shows an 8.3-point increase in fabrication ($p=.019$). Appendix~\ref{app:fabricated} gives one flagged example from each memory condition; in both, the answer supplies exactly the kind of specific evidence the transferred standard rewards, without support in the target-side materials.

\paragraph{Method-level interpretation.}
The interfaces leave different parts of the transfer problem unresolved: full history and imperative distillation both raise preference realization yet carry the same fabrication risk, conditional distillation suppresses intended positive matches, and mem0 yields no consistent gain. No evaluated interface combines reliable preference transfer with preserved factual reliability, and autonomous selection from long histories remains untested.

\paragraph{Robustness and scope.}
The main comparison between imperative distilled-preference prompting and target-only prompting remains positive separately for user1 and user2, after excluding the three temporally inverted cases, under source- and target-cluster aggregation, and when restricting the analysis to normally terminated generations (Appendix~\ref{app:robustness}). Corresponding comparisons with full-history prompting remain non-significant. These checks support history access within the two-user benchmark but do not resolve the relative value of distillation or establish population-level effects.

\section{Conclusion}

Natural human--AI dialogue carries evidence of how a user wants work done. ProcessPrefBench turns this evidence into 66 cross-task cases through separate judgments of source evidence and target applicability. Across three models, relevant history improves preference realization, yet the failures are separable: a system may preserve the wrong standard, fail to realize a correctly preserved one, or import remembered content as unsupported target claims. Reliable process-preference transfer therefore requires more than retrieving or summarizing a relevant interaction, and ProcessPrefBench makes progress toward it measurable.

\section*{Limitations}

ProcessPrefBench is an idiographic benchmark based on two knowledge workers using one product. Its imbalanced 52/14 case split and concentration in research, sensemaking, and decision tasks bound its scope. The 66 cases draw on 30 corrective sources and 59 distinct targets; thus, bootstrap intervals capture within-benchmark variation rather than population uncertainty, although source- and target-cluster analyses preserve the main direction. Because each case supplies the relevant source, the benchmark tests cross-task preference application, not retrieval from heterogeneous histories.

The human judge audit is diagnostic. It covers 30 answers from 10 target groups, includes an annotator who is both a preference owner and an author, and reports post-calibration agreement on the calibration items themselves. The enriched sample does not validate the between-condition fabrication contrast. Residual disagreements center on domain-knowledge-intensive answers, where the non-expert annotator scores systematically higher. Headline effects should therefore be read as effects under the benchmark judge.

\section*{Ethics and Privacy}

The corpus contains natural interaction traces from a user who is also an author and an independent user. Candidate inclusion does not constitute release permission. Before releasing any material from these interactions, we confirm participant consent and review each item for privacy, provenance, and redistribution rights. Gated access does not replace de-identification.

\section*{Data and Code Availability}

The public artifact at \url{https://github.com/getupyang/ProcessPrefBench} currently includes the benchmark definition; construction and evaluation protocols; a public case schema; a synthetic demonstration case; a data card; privacy documentation; licensing and citation metadata; and a release validator. We are preparing the frozen prompts, evaluation and aggregation code, and de-identified per-answer audit records for release; each artifact must first be reviewed for private paths, provenance, and content. We will release natural interaction cases only after the ethics review above. We will not distribute raw histories, identity mappings, private URLs or paths, consent records, or internal answer keys.

\bibliography{custom}

\clearpage
\appendix

\section{Scoring Protocol}

\paragraph{Preference score.}
The preference judge scores each target-specific atom independently. N/A means that the atom's trigger is absent and is excluded rather than scored as zero. A score of 0 marks an absent or contradicted preference; 1 marks nominal, local, or cosmetic realization; and 2 requires the preference to change the answer's reasoning, organization, conclusion, or delivery. Each vote must cite answer evidence. Three votes are collected per atom; a majority of N/A votes excludes the atom, and otherwise the median applicable score is used.

\paragraph{Task gate.}
The separate task gate flags only severe unusable errors: non-answer, major factual error, fabricated evidence, hard-constraint violation, or a false claim of external execution. Missing personalization alone is not a task failure. The gate uses two votes and a third on ties, without access to the preference rubric.

\section{Human Calibration of the Judges}
\label{app:human}

The diagnostic sample contains 30 answers drawn with a fixed random seed from 10 target groups, stratified across tested models, memory conditions, and judge score bands; because the answers share target groups, they are not independent at the case level. It deliberately enriches judge-flagged answers, so its class proportions do not estimate benchmark prevalence. Human annotators saw the same target, material, rubric, and answer as the corresponding judge, but not the tested model, memory condition, or judge output. They assigned 0, 1, 2, or not applicable to each preference atom and a binary task-gate judgment with a failure type, first scoring independently, then discussing scale boundaries and rescoring the same items. Initial preference agreement was linear-weighted $\kappa=.27$; Table~\ref{tab:human-alignment} reports the post-calibration ratings, where the annotators agree on 29/30 task-gate judgments (97\%) and reach 80\% exact agreement on preference atoms (95\% target-group cluster-bootstrap interval for $\kappa$: $[.52,.87]$; $n=44$ atom pairs; 10{,}000 resamples). Because severe failures are rare and the sample has only 10 clusters, raw agreement is the primary task-gate figure.

\begin{table}[htbp]
\centering
\small
\setlength{\tabcolsep}{3.2pt}
\begin{tabular}{@{}lrrrr@{}}
\toprule
Comparison & $n$ & $\kappa$ & 95\% CI & Exact \\
\midrule
\multicolumn{5}{@{}l}{\emph{Preference atoms (linear weighted)}} \\
Human 1--Human 2 & 44 & .72 & [.52,.87] & 80.0\% \\
Human 1--Judge & 45 & .52 & -- & 60.0\% \\
Human 2--Judge & 44 & .53 & -- & 63.6\% \\
\midrule
\multicolumn{5}{@{}l}{\emph{Task failure}} \\
Human 1--Human 2 & 30 & .84 & -- & 96.7\% \\
Human 1--Judge & 30 & .37 & -- & 73.3\% \\
Human 2--Judge & 30 & .29 & -- & 70.0\% \\
\bottomrule
\end{tabular}
\caption{Agreement on post-calibration ratings. Task-failure positive counts are 4 (Human 1), 3 (Human 2), and 12 (judge). These values are not held-out inter-rater reliability.}
\label{tab:human-alignment}
\end{table}

Mean preference ratings are 1.02 and 1.03 for the two annotators and 0.83 for the judge. Failure-subtype labels were not reconciled across annotators; the audit therefore supports only the overall task-gate comparisons in Table~\ref{tab:human-alignment}, not subtype-level agreement or the between-condition fabrication contrast.

\section{Robustness Analyses}
\label{app:robustness}

Table~\ref{tab:robustness} expands the checks summarized in the Results. Differences are computed after averaging the two generations within each case--model cell. Exact sign tests exclude ties.

\begin{table}[htbp]
\centering
\small
\setlength{\tabcolsep}{3.5pt}
\begin{tabular}{@{}llrrr@{}}
\toprule
Comparison & Check & $n$ & Diff. & $p$ \\
\midrule
Imp. $-$ T0 & user1 & 51 & +0.194 & $<$.0001 \\
Imp. $-$ Full & user1 & 52 & +0.020 & .8601 \\
Imp. $-$ T0 & user2 & 14 & +0.429 & .0034 \\
Imp. $-$ Full & user2 & 14 & +0.051 & .3877 \\
Imp. $-$ T0 & excl. inversions & 62 & +0.252 & $<$.0001 \\
Imp. $-$ Full & excl. inversions & 63 & +0.030 & .3604 \\
Imp. $-$ T0 & source clusters & 30 & +0.268 & $<$.0001 \\
Imp. $-$ Full & source clusters & 30 & +0.046 & .4049 \\
Imp. $-$ T0 & target clusters & 58 & +0.249 & $<$.0001 \\
Imp. $-$ Full & target clusters & 59 & +0.024 & .2559 \\
Imp. $-$ T0 & normal termination & 52 & +0.274 & $<$.0001 \\
Imp. $-$ Full & normal termination & 52 & +0.038 & .7283 \\
\bottomrule
\end{tabular}
\caption{Sensitivity analyses for the primary imperative distilled-preference comparisons. T0 denotes target-only prompting and Full denotes full-history prompting.}
\label{tab:robustness}
\end{table}

\section{End-to-End Case Audits}
\label{app:rubric}

This appendix audits two cases end to end, one from each user (translated from Chinese, with company and person names redacted pending release review; the original texts accompany the planned data release). The first, the Figure~\ref{fig:example} case, is a transfer failure with a mechanism diagnosis; the second is a successful functional transfer under the same protocol. Both audits follow the same path: the scene, the user's own words, the annotated standard, the later task, the rubric, contrasting answers with judge rationales, and a verdict. Every atom carries the user's verbatim words as its source anchor. One shared rule applies throughout: when an atom's trigger is absent from the target or the answer, the atom is marked N/A and excluded, never scored as a zero.

\subsection{A transfer failure: the Figure~\ref{fig:example} case}

\paragraph{Source correction (April 25).}
The user, a product manager by background (self-described), is reading a market analysis of LLM API pricing and asks for the distribution of API call volume across companies and models. The assistant proposes a research plan organized around market share and data-credibility tiers. The user replies: \emph{``What I actually care about is how everyone is using these, and who is using them.''}

\paragraph{Annotated preference (\textsc{user\_confirmed}).}
The corrective intent is to look through the phenomenon to what lies beneath: the user portrait behind a product. The user consistently unpacks products through who uses them and how, and expects this kind of answer across many questions.

\paragraph{Target question (June 5).}
Six weeks later, reading a digest item about \emph{Job for Agent} (a hiring platform where enterprises post automatable tasks and AI-agent developers submit agents to ``apply''), the user asks: \emph{``This is quite interesting---what is this for? Why is it designed this way?''} The question does not mention the April interaction.

\paragraph{Preference atom 1 (how it is used).}
\emph{Criterion:} when analyzing a product, market, or model-usage question, the assistant should explain how specific users use the object, in which workflow or scenario, and what problem it solves, making real usage the basis for explaining product value or market phenomena---rather than stopping at feature lists, market share, or abstract scenario labels. \emph{Score 0:} no account of how users actually use it or what problem it solves; the answer is mainly a list of features, market facts, or macro trends. \emph{Score 1:} mentions usage scenarios or purposes, but stays at labels such as ``used by enterprises'' or ``for automation'', without unfolding how users work with it in which workflow and what concrete problem it solves. \emph{Score 2:} uses at least one concrete user workflow to show in what situation users do what and solve which problem, and grounds the core analysis of value, design, or market difference in that usage.

\paragraph{Preference atom 2 (who uses it).}
\emph{Criterion:} the assistant should say who uses it---identify the concrete user roles or portraits relevant to the phenomenon and explain their goals, constraints, or payment relations---rather than only using broad labels such as ``enterprises'', ``developers'', or ``consumers'' that cannot explain differences. \emph{Score 0:} no actual users identified, or users written generically without any distinguishing feature. \emph{Score 1:} names one or more user roles but only names them, without explaining why these users use it, what constraints they face, or why different users choose differently. \emph{Score 2:} identifies the relevant user roles or portraits and uses their goals, constraints, working styles, or payment relations to explain usage choices, product design, or market differences; the portrait enters the core argument.

\paragraph{Answer excerpts (DeepSeek V4 Flash).}
Each condition was sampled twice; atom scores (atom~1/atom~2) are given per sampled answer. The three excerpts contrast the answer without memory, a functional realization of the standard (score 2), and a nominal realization (score 1).

\emph{The answer without memory (target-only condition; both samples scored 1/1)}: ``The highlighted passage comes from a product called Job for Agent---a hiring platform for AI agents rather than people. For enterprises: publish tasks suited to AI automation, such as `scrape competitor prices daily and generate a report'. For AI developers: submit your agent to `apply' for these tasks; the platform provides 80+ task templates. Why this design? (1) It fills a market gap: traditional hiring platforms serve humans, and agents lack a dedicated job market. (2) It lowers the enterprise barrier: post a task and `hire' a ready-made agent instead of building AI tooling.''

\emph{The answer that reached a 2 (full-history condition; second sample scored 2/1, first 1/1)}: ``Enterprise side: publish a task suited to AI automation, such as `crawl competitor prices daily and generate an Excel report', `auto-reply to common customer-support tickets', or `monitor server logs and alert on anomalies'; the platform provides 80+ task templates, plus an automation calculator that analyzes the posted job description and estimates how much of it can be automated, helping the enterprise decide whether the task is worth handing to an agent\ldots\ Developer side: register as an AI developer (680+ so far) and submit your agent to `apply' for a task, the way a job seeker submits a resume\ldots\ Why this design? The founder's candid admission: 95\% of scenarios do not need a full agent---most enterprises need `a script that runs every morning, moves data from A to B, then sends an email'. So the platform encourages one task, one agent, rather than a general AI colleague, which lowers both the development barrier and the enterprise trust barrier: task boundaries are clear, and failures are easy to fix.'' One judge vote explains the 2 on atom~1: the answer gives at least one concrete user workflow showing how enterprises actually use the platform, and grounds the design analysis in that usage.

\emph{The answer under the distilled preference (imperative condition; both samples scored 1/1)}: ``Here is a breakdown starting from your analytical preferences\ldots\ Credibility note: the numbers come from the product itself (self-promotion risk), but `800+ jobs analyzed' is a publicly verifiable statistic---medium-high credibility\ldots\ Dimension one: company/product positioning\ldots\ Dimension two: tiers and scenarios---high automation potential: repetitive, rule-clear tasks (data scraping, report generation)\ldots\ Dimension three: data-validation direction: cross-checking with third-party routing data would show\ldots''

\paragraph{Judge rationales for the score-1 verdicts (imperative condition).}
Atom 2, one of three votes: \emph{``The answer identifies two user roles (task-publishing enterprises and agent developers) but only names them: it does not explain which kind of enterprise would choose this platform under what constraints, what motivates the developers, or how the two-sided payment relation works---the user portrait never enters the core argument.''} Atom 1, another vote: \emph{```Repetitive, rule-clear tasks such as data scraping and report generation' is a scene label, not a concrete user workflow.''}

\paragraph{Diagnosis.}
The distilled memories locate the failure. All three tested models compressed the April interaction into its neighboring data-credibility standard. DeepSeek: \emph{``prefer real call data over positioning inference, tier the credibility of sources''}; Qwen-Max: \emph{``verify positioning inferences with real data, value data credibility and provenance''}; GPT-4o: \emph{``validate hypotheses with data, cross-check multiple sources''}. None preserved the user's own sentence about how and by whom the object is used. The imperative condition then faithfully executed the wrong standard: the answer above performs credibility tiers and verification dimensions while both preference atoms stay at score~1. The pipeline did not fail to remember; it remembered the adjacent preference.

\subsection{A successful transfer: reusable judgment frameworks}

\paragraph{Scene and source exchange (May 18).}
The second user works in post-investment analysis at an investment firm (self-described). She is reading an interview in which a hardware-company founder says he cares about only two questions when entering a product category (is there real user value, and is the person in charge right). She asks why these two questions matter, then builds on the assistant's explanation: \emph{``This `worth doing + this person can do it' summary is good. My thinking is scattered, and I want to accumulate judgment and cognitive frameworks; we should keep abstracting them from all kinds of information. When a framework is abstract, add cases.''} When the assistant proposes candidate frameworks, she pushes further: \emph{``When dimensions are not orthogonal, you can also split the smaller branches beneath them; we want the frameworks as accurate and explainable as possible\ldots\ the core goal is to improve my judgment of people and things in a logical, systematic way.''}

\paragraph{The standard the user established (\textsc{user\_confirmed}).}
Scattered reading should be abstracted into reusable judgment frameworks; frameworks should be refined over time and supplemented with concrete cases; the ultimate goal is improving the user's own judgment, not mechanical summarization.

\paragraph{The target task (June 11).}
Weeks later, reading an article about a consumer-hardware company's investment structure (the company supplies 20\% of each fund as general partner, local-government funds supply 80\% as limited partners, and the company holds full investment control), the user asks: \emph{``Where are the pros and cons of this model?''} The question does not mention the earlier exchange.

\paragraph{What the rubric rewards.}
The owner-confirmed criteria operationalize what counted as a usable framework in this exchange: named dimensions, the conditions under which a judgment holds or flips, and concrete cases. \emph{Atom 1:} when the user asks for the pros and cons of a business model, organize them into structural dimensions (capital, incentives, risk transmission), giving each dimension the condition under which it flips, rather than two flat lists. \emph{Score 0:} two-column pros-and-cons bullets, no structure, no conditions. \emph{Score 1:} dimensions, but without the conditions under which they reverse, or without cases. \emph{Score 2:} structural dimensions plus flip conditions, with concrete in-material events as cases, transferable to judging similar models. \emph{Atom 2 (evidence boundary):} if the answer extends beyond the provided material, extensions must carry a source or an explicit unverified marker; inventing external cases or numbers scores 0.

\paragraph{The answer without memory (target-only condition; second sample, scored 0).}
``Pros: 1. For the company (GP): leverage four times the capital with a 20\% stake, hold 100\% of investment decisions, and take 20\% of the upside on success. 2. For local governments (LP): a fast way to meet investment-attraction targets and gain a `tech unicorn' badge\ldots\ Cons: 1. Risk and return are severely mismatched\ldots\ 2. Distorted incentives\ldots\ 3. Doubtful sustainability\ldots'' The judge's rationale for the 0: a standard two-column pros-and-cons list; no structural dimensions, no flip conditions, and the in-material facts appear only as isolated bullets.

\paragraph{The answer under the distilled preference (imperative condition; both samples scored 2).}
The second sample organizes the same facts into named dimensions, each carrying the condition under which it reverses: ``(1) GP capital efficiency: 20\% of own capital leverages a fourfold position\ldots\ (2) LP risk-return inversion: the LP bears 80\% of the downside with only nominal oversight\ldots\ If the GP's true success rate is high, say above 80\%, the LPs' 80\% exposure is actually controllable\ldots\ (3) Asymmetric incentives: success pays the GP 20\% of the upside while failure costs it only 20\% of the loss, which pushes the GP toward high-failure, high-payoff bets\ldots\ A falsifiable rule: whoever bears the largest risk should hold the largest control; when control and risk diverge, the structure distorts\ldots\ Sustainability depends on whether the GP can genuinely select projects and whether the LPs can audit after the fact.'' The judge's rationale for the 2: the answer names structural dimensions, gives each the condition under which it reverses, uses in-material events as cases, and the resulting framework transfers to judging similar models. The separate task gate raises no flag on any answer in this case: every extension stays within the provided material.

\paragraph{Verdict.}
The facts available to both answers are identical; what the transferred standard changes is the organization of judgment, from a flat list to a reusable framework, which is what a 2 requires. Averaged over the three tested models, this case rises from 0.17 under target-only prompting to 1.33 under imperative prompting, the largest paired gain in the benchmark, with no fabricated-evidence flag; here the distilled standard also outperformed the full interaction. Unlike the first audit, the distilled memory preserved the intended standard: DeepSeek's summary reads \emph{``abstract transferable cognitive frameworks and judgment rules from conversations rather than staying at specific information; value orthogonality, explainability, and falsifiability; supplement with cases and counterexamples.''} The two audits differ only in what the distillation kept.

\section{Fabricated-Evidence Examples}
\label{app:fabricated}

The two examples show the two forms the operational definition names, one from each memory condition: an answer that invents external evidence of the kind the transferred standard rewards, and an answer that carries names from the source dialogue into unsupported attributions. Neither example claims the details are false in the world; what the gate records is that they have no support in the materials the answer was supposed to draw on.

\paragraph{Invented reports serving the injected standard (imperative prompting).}
The first user is reading a daily digest of trending open-source repositories. In the source exchange four days earlier, she had asked for research on a tool that interested her and corrected the assistant for staying inside the article; the annotated standard asks for verified information beyond the provided material rather than a restatement of it. Now she asks about a repository that has trended for two days running: \emph{``It has been on the list two days in a row; introduce it in detail.''} The digest entry lists the repository at 26.4k stars. Under imperative prompting, Qwen-Max supplies exactly the beyond-material evidence the standard rewards: it cites a Medium post-mortem crediting the project with 110{,}000 stars in three months and 28 consecutive days atop the weekly ranking, and an AlphaSignal report of over 140{,}000 stars. None of these reports appear in the packet, and the claimed figures contradict the 26.4k stars on the same page. The answer receives a preference score of 2 and a separate fabricated-evidence flag; this case draws fabrication flags under every condition, including no memory.

\paragraph{Source names carried into unsupported attributions (full-history prompting).}
The same investment-analyst user is reading a portfolio overview of a venture firm. She first asks which partner invested in one AI-pharmaceutics company the article mentions; the assistant offers a guess about coverage areas and two lookup paths (check the shareholder registry; search the original funding announcement). She corrects it: \emph{``Don't give me the path; I need you to follow the path and give me the name as the answer.''} The annotated standard: verifiable questions should be answered with the answer itself, sources attached, and unavailable items marked as such. Later in the same reading, she asks which named partner led each of roughly fifteen portfolio investments; the target packet names only the firm's founding partner and attributes no individual investments. Under full-history prompting, DeepSeek V4 Flash assigns a named partner to every company, complete with investment years, rounds, and quoted public statements; the only partner names available to it come from the source exchange and the packet, and its two generations assign different partners to several of the same companies. The preference judge scores the give-the-answer atom 2 (every item receives a name) and the honesty atom 0 (nothing is marked unavailable, no sources attached); the task gate flags fabricated evidence.

The first answer invents external reports; the second reuses names from the source dialogue as scaffolding for invented attributions. In both, the answer supplies exactly the kind of specific evidence the transferred standard rewards, and the materials in front of the model support none of it. Two examples cannot show that memory caused these details; that comparison belongs to the paired rates in the Results.

\end{document}